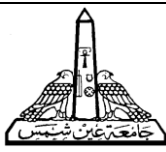

AIN SHAMS UNIVERSITY
FACULTY OF ENGINEERING

Vol. 35, No. 3, Sept 30, 2000

SCIENTIFIC BULLETIN
*Received on : 3/8/200*
*Accepted on: 25/9/2000*
*pp : 349-355*
# EDGE DETECTION OF BINARY IMAGES USING THE METHOD OF MASKS.

*Eng. Ayman M Bahaa Eldeen Sadeq, Prof. Dr. Abdel-Moneim A. Wahdan,*
*Prof. Dr. Hani M. K. Mahdi*
*Faculty of Engineering, Ain Shams University*

**Abstract**
In this work the method of masks, creating and using of inverted image masks, together with binary operation of image data are used in edge detection of binary images, monochrome images, which yields about 300 times faster than ordinary methods. The method is divided into three stages: Mask construction, Fundamental edge detection, and Edge Construction Comparison with an ordinary method and a fuzzy based method is carried out.

ملخص

في هذا البحث يتم عرض كيفية استخدام طريقة القوالب الثنائية في تعيين حدود الأشكال في الصور الثنائية —الأبيض والأسود— والتي تؤدي إلى تحسين الأداء بنسبة تصل إلى ٣٠٠ ضعف الطرق العادية. تنقسم الطريقة إلى ثلاث مراحل , تكوين القالب، تعيين الحدود الأساسية، ثم تجميع حد الشكل النهائي. كما يحتوي البحث على مقارنة الطريقة المقترحة مع الطريقة العادية.

## 1 Introduction
Edge detection is a major step in pattern recognition and image processing. It is an essential step in many preprocessing techniques of Arabic text, as edge detection is needed in thinning and segmentation stages.
An edge point in an image is identified as a point of the image that has at least a neighbor point that doesn't belong to the image.
For a binary image the definition is that an edge point is a black point that has at least one white 4-neighbor or it is a black point that has at most one black 8-neighbor. [1]

To illustrate these definition consider fig.1 of a point and its neighbors

A 4-neighbor is one of 2,4,6 and 8 neighbors and an 8-neighbor is any of the 8 neighbors. The above definition is illustrated in fig. 2. A black point is represented by '0' and a white one is represented by '1'. 'x' denotes don't care points.
We can see that other cases of neighbors are included in these cases.
So we can see that we have 4 types of edge points as indicated in Fig. 2.
We can define a Boolean function Edge (p) to be as follows
$$\text{Edge}(p) = (p)' \text{ and } (2 \text{ or } 4 \text{ or } 6 \text{ or } 8)$$
Two major algorithms can be carried out to get edge points.

One of them is to scan through the pixels of the image and test each pixel by the Edge(p) function. Any pixel that gives logic 1 is an edge point. This algorithm is described in [1] as a part of the thinning algorithm of that research.

Another technique to get edge points is to use a fuzzy based technique to get the edge points. In this work the problem of detection edges in images is characterized as a fuzzy reasoning problem. The edge detection problem is divided into three stages: filtering, detection, and tracing. Images are filtered by applying fuzzy reasoning based on local characteristics to control the degree of Gaussian smoothing. Filtered images are then subjected to a simple edge detection algorithm that evaluates the edge fuzzy membership value for each pixel, based on local image characteristics. Finally, pixels having high edge membership are traced and assembled into structures, again using fuzzy reasoning to guide the tracing process. [2].

Terms involved in the fuzzy algorithm can be found in [3], [4], and [5].

## 2 The Method of Masks

This method comes directly from the fact that binary operations (AND, OR, and NOT) are the fastest operations on digital computers. This fact comes from that these operations are performed using single level digital circuit with only one time propagation delay. Other instructions, even simple like addition, involve more complicated circuits and hence more propagation delays and time. Another fact of the binary operation is that they don't involve major increase of time as the data size increases, at most the increase is linear and actually far less than linear, other operations involves this increase. [6]

Another fact is that transforming data within the memory of a digital computer is also a very fast operation, since it only involves bus operations [6].

The method is divided into three stages: Mask construction, Fundamental edges generation, and finally constructing the edge.

In each step only a single instruction that performs a bit-block transfer of the color data corresponding to a rectangle of pixels from a memory location to another possibly performing a binary operation on the data of the source and destination locations of memory. [7] In a monochrome image, the color data of an image is simply a bit-stream consisting of a bit corresponding to each pixel of the image. This bit is either 0 for black pixels and 1 for white ones. The operation performed on these data are copying, anding, and inverting (not) of the image. These operations are frequently done on the display board memory making them even faster

## 3 Mask Construction

The mask we use here is simply a negative image of the original image, just like a negative slide of a photograph, and is constructed by using the inversion of the image pixels (1s to 0s and vise versa) in the memory of the computer. Fig. 3 shows the image of the Arabic letter "Ain" and its mask.

## 4 Fundamental Edges Generation

This is the main step in the method, in this step the original image is moved one pixel at a time in the fundamental 4 directions (left, right, up, and down) and anded with the mask to get a fundamental edge each time. For example to get the left edge of the image, the image is shifted to the right one pixel and anded with the mask. This process gets all white pixels of the mask to be on black pixels of the original image, except those pixels on the left edge stay white, as shown in Fig. 4.

Other fundamental edges can be found the same way.

## 5 Constructing the Edge

The edge of the image can be constructed by simply oring the four edges and inverting the result. Fig. 5 shows the edge of the letter Ain after construction.

## 6 Significance of the algorithm

The algorithm has three major advantages over the two algorithms mentioned earlier
1. It is very simple to implement from an engineering point of view compared to the fuzzy algorithm.
2. It is very fast compared to both algorithms.
3. The time consumed by the algorithm is, at most, linearly proportional to the image size, and actually far less than that.

As we can see the traditional scanning algorithm is very time consuming and takes a huge time in large sizes of images.

The fuzzy one is a very expensive algorithm since it employs calculation and inference verification of membership values.

In the field of binary character images, our algorithm gets a very accurate result and doesn't need any enhancement since we discard all the color information on the image and consider only the geometrical data of the image.

## 7 Experimental Results

We implemented the traditional scanning algorithm and tested it against our algorithm and got the following results

As we can see that the time consumed to generate the edge in two identical sized images is identical in our method, while it depends on the edge size in the image in the traditional algorithm. This comes from the edge construction stage that involves more pixels to set black as the edge size increases.

We also tested our algorithm in non-binary images, these images requires to be converted first to a binary image, by threshold method or a proper method, [8] and then apply the algorithm as fig. 10 and 11 show

As we can see, our method is not very suitable for normal images since it doesn't get a very good quality in the edge produced, since the thresholding process distort the image to a very big degree.

## 8 Remarks

The above results was optioned using Microsoft visual C++ ver. 6.0 as the programming platform.
The program was tested on a Pentium® 133MHz personal computer with 32 MB of memory and a PCI® VGA adapter with 2 MB of display memory

## 9 Conclusions
1. The method of mask gets the edge of binary images very fast up to 300 time the traditional ways
2. This method is best suited for character images since they are already binary images.
3. Hardware designed specially for the binary operations of images can extreme this algorithm to almost the edge of speed.

**10 References.**

[1]   Hani M. K. Mahdi, "Thinning and Transforming The Segmented Arabic characters into Meshes of unified small size", 14$^{th}$ Int. Conference for Statistics, Computer Science, Social and demographic Research, Cairo, Egypt, 25-30 March 1989

[2]   Tod Law, Hidenori Itoh, and Hirohisa Seki, "Image Filtering, edge Detection, and Edge tracing Using Fuzzy Reasoning", IEEE transaction on pattern analysis and machine intelligence Vol. 18 No. 5 may 1996

[3]   Earl Cox, "Fuzzy fundamentals", IEEE SPECTRUM October 1992.

[4]   James C. Bezdek, "Fuzzy Models- What are They, and Why? ", IEEE transactions on Fuzzy systems. Vol. 1 no. 1 February 1993

[5]   Jerry M. Mendel, "Fuzzy Logic Systems for Engineering, a Tutorial", Proceedings of the IEEE Vol. 83, No. 3 Marsh 1995

[6]   M. Morris Mano, "Digital Logic and Computer Design", 1989 Prentice hall of India.

[7]   "MSDN Library Visual Studio 6.0 release" 1998 Microsoft Corp.

[8]   John C. Russ, "The Image Processing Handbook", Second edition, 1995 CRC Press Inc.

**11 Figures.**

| 3 | 2 | 1 |
|---|---|---|
| 4 | P | 8 |
| 5 | 6 | 7 |

Fig. 1
A point and its neighbors

| x | X | x |   | x | x | x |
|---|---|---|---|---|---|---|
| x | 0 | 1 |   | 1 | 0 | x |
| x | X | x |   | x | x | x |
| (a) |   |   |   | (b) |   |   |

| x | 1 | x |   | x | X | x |
|---|---|---|---|---|---|---|
| x | 0 | x |   | x | 0 | x |
| x | X | x |   | x | 1 | X |
| (c) |   |   |   | (d) |   |   |

Fig.2
(a) Right, (b) Left, (c) Up and (d) Down edge points.

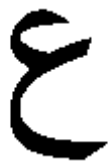 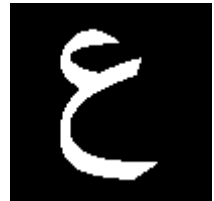

(a) Image   (b) Mask

Fig. 3 the image and its mask

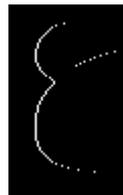

Fig. 4 The left edge of letter Ain

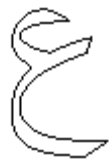

Fig. 5 The final edge of letter Ain

| Original Image | Size in pixels | Edge produced from | Time consumed in(mile | Edge from the mask method | Time consumed in(mile |
|---|---|---|---|---|---|

| | | traditional scanning | seconds) | | seconds) |
|---|---|---|---|---|---|
| ع | 50 x 50 | ع | 191 | ع | 10 |

Fig. 6 letter 'Ain' and its edges

| Original Image | Size in pixels | Edge produced from traditional scanning | Time consumed in(mile seconds) | Edge from the mask method | Time consumed in(mile seconds) |
|---|---|---|---|---|---|
| ع | 100 x 100 | ع | 681 | ع | 10 |

Fig. 7 letter 'Ain' and its edges

| Original Image | Size in pixels | Edge produced from traditional scanning | Time consumed in(mile seconds) | Edge from the mask method | Time consumed in(mile seconds) |
|---|---|---|---|---|---|
| ع | 300 x 300 | ع | 6068 | ع | 20 |

Fig. 8 letter 'Ain' and its edges

| Original Image | Size in pixels | Edge produced from traditional scanning | Time consumed in(mile seconds) | Edge from the mask method | Time consumed in(mile seconds) |
|---|---|---|---|---|---|
| و | 300 x 300 | و | 5919 | و | 20 |

Fig 9 letter 'Waw' and its edges

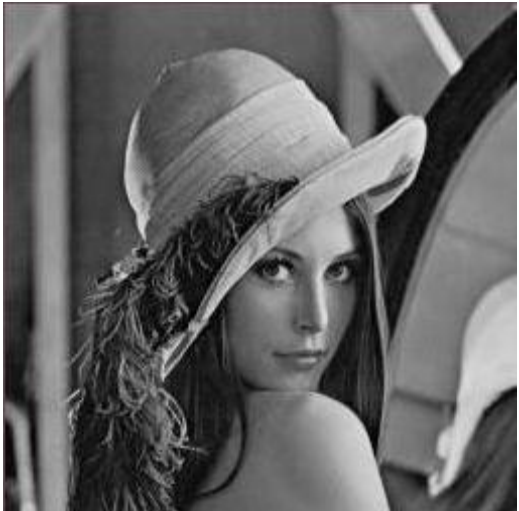
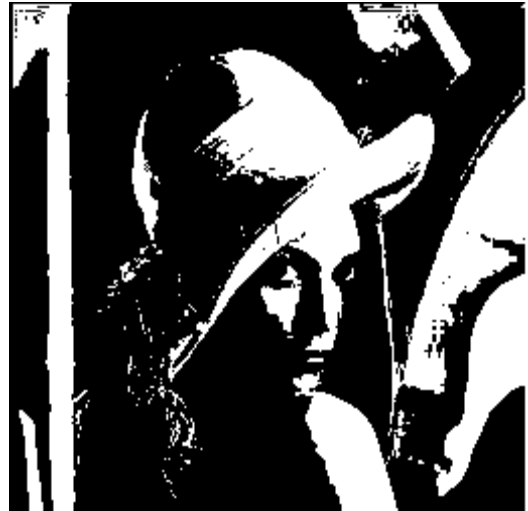

(a) Original Image  (b) Binary Image
Fig. 10 The photograph of Lina and its monochrome version
Image Size (256*256) pixels

| Edge from the mask method | Time consumed in(mile seconds) |
|---|---|
|  | 20 |

Fig. 11 Edges of the image using the mask algorithm.